\documentclass[Afour,sagev,times]{sagej}

\makeatletter
\NAT@superfalse
\makeatother
\setcitestyle{numbers,square,comma,sort&compress}

\usepackage{moreverb,url}
\usepackage[colorlinks,bookmarksopen,bookmarksnumbered,citecolor=red,urlcolor=red]{hyperref}
\newcommand\BibTeX{{\rmfamily B\kern-.05em \textsc{i\kern-.025em b}\kern-.08em
T\kern-.1667em\lower.7ex\hbox{E}\kern-.125emX}}
\usepackage{color}
\usepackage{graphicx}
\usepackage{caption}
\usepackage{mathtools}
\usepackage{url}
\usepackage{graphicx} 
\usepackage{float}
\usepackage{caption}
\usepackage{subcaption}
\usepackage{textcomp}
\usepackage{xcolor}
\usepackage{amssymb,amsfonts}
\usepackage{amsmath}
\usepackage{tikz}
\usetikzlibrary{shapes.geometric, arrows, positioning}
\usetikzlibrary{arrows.meta, positioning, shadows.blur}
\usepackage{multirow}

\newcommand{\bm}{\boldsymbol}

\def\volumeyear{2025}
\def\journalname{International Journal of Engine Research}

\begin{document}

\bibliographystyle{SageV} % Vancouver-style numeric citation format

\runninghead{Zinage, et al.}

\title{A Causal Graph-Enhanced Gaussian Process Regression for Modeling Engine-out NOx}

\author{Shrenik Zinage\affilnum{1}, Ilias Bilionis\affilnum{1}, Peter Meckl\affilnum{1}}

\affiliation{\affilnum{1}School of Mechanical Engineering, Purdue University, West Lafayette, IN, USA}

\corrauth{Shrenik Zinage}

\email{szinage@purdue.edu}

\begin{abstract}
The stringent regulatory requirements on nitrogen oxides (NOx) emissions from diesel compression ignition engines require accurate and reliable models for real time monitoring and diagnostics. Although traditional methods such as physical sensors and virtual engine control module (ECM) sensors provide essential data, they are only used for estimation. Ubiquitous literature primarily focuses on deterministic models with little emphasis on capturing the various uncertainties. The lack of probabilistic frameworks restricts the applicability of these models for robust diagnostics. The objective of this paper is to develop and validate a probabilistic model to predict engine-out NOx emissions using Gaussian process regression. Our approach is as follows. We employ three variants of Gaussian process models: the first with a standard radial basis function kernel with input window, the second incorporating a deep kernel using convolutional neural networks to capture temporal dependencies, and the third enriching the deep kernel with a causal graph derived via graph convolutional networks. The causal graph embeds physics knowledge into the learning process. All models are compared against a virtual ECM sensor using both quantitative and qualitative metrics. We conclude that our model provides an improvement in predictive performance when using an input window and a deep kernel structure. Even more compelling is the further enhancement achieved by the incorporation of a causal graph into the deep kernel. These findings are corroborated across different verification and validation datasets.
\end{abstract}


\maketitle

\section{Introduction}
\label{sec:intro}

Given the detrimental impact of nitrogen oxides (NOx) on environmental and human health, stringent regulations~\citep{epa} are essential to mitigate these effects and ensure sustainable urban air quality. In this context, developing robust predictive models is crucial for guiding the evolution of internal combustion engine technology, allowing manufacturers to comply with evolving global environmental legislation and effectively reducing air pollutants, including those contributing to greenhouse gas emissions. Research on predictive models for engine-out NOx emissions remains a dynamic and ongoing area of focus within the diesel powertrain research community.
Although extensive research in this field is documented in the open literature~\citep{aliramezani2022modeling}, more studies need to be conducted, particularly in analyzing uncertainties associated with these models. Typically, such models are intended for real time use, with their predictions commonly used for control or diagnostic applications. Therefore, it is critical to develop probabilistic models. These models should not only predict emissions but also clearly indicate the expected range of errors or provide a distribution of possible prediction errors.

Given the extensive research on modeling engine-out NOx, various methods have been explored to address the complexities and uncertainties inherent in these systems. Early research developed physics based models to describe the chemical and thermal mechanisms that produce NOx in diesel engines. For example,~\cite{asprion2013fast} presented a fast physics based approach for estimating NOx levels. Another study~\cite{aithal2010modeling} used chemical kinetics with finite reaction rates to simulate NOx formation, demonstrating how combustion dynamics impact emission levels. Although these approaches provide deep understanding, they often demand significant computational resources and require precise knowledge of engine conditions.
In recent years, data driven techniques have gained traction for NOx emission prediction, as these models can learn patterns directly from observed data. For instance,~\cite{fang2022artificial} applied artificial neural networks to estimate transient NOx outputs in high speed diesel engines. In another case,~\cite{shin2020deep} combined deep learning with Bayesian hyperparameter optimization to forecast NOx emissions during rapidly changing engine operations. Other researchers have turned to machine learning algorithms such as support vector machines and used optimization techniques like particle swarm optimization to model emissions in homogeneous charge compression ignition engines~\citep{gordon2023support}.
Hybrid modeling strategies that bring together physical principles and machine learning have also been investigated. For example,~\cite{shahpouri2021soot} explored machine learning methods that incorporate physical insights to predict soot emissions. However, these deterministic models often overlook the uncertainties associated with sensor data and engine dynamics. To address this limitation, probabilistic frameworks have been proposed. ~\cite{yousefian2021bayesian} used Bayesian inference techniques along with uncertainty quantification methods specifically targeting hydrogen enriched and lean premixed combustion processes. Similarly, ~\cite{cho2018structured} introduced a systematic methodology for performing uncertainty analysis in predictive models focused on engine-out NOx, highlighting the need for robust uncertainty quantification.

Gaussian processes (GPs)~\citep{williams2006gaussian} are popular tools in Bayesian analysis because they offer clear interpretations and robust ways to measure uncertainty. GPs usually depend on a limited set of kernel parameters, which are adjusted to maximize the marginal likelihood. Most practical implementations, however, use a predetermined kernel, which restricts the model’s capacity to learn complex patterns from the data. As a result, GPs often serve as smoothers and may not perform well with data that has intricate structures or lies in high dimensional spaces.

On the other hand, deep neural networks (DNNs)~\citep{lecun2015deep} have shown exceptional ability to learn useful representations, making them highly effective at predicting new data. Yet, traditional neural networks are usually deterministic, which can lead to overconfident outputs~\citep{guo2017calibration} and unreliable uncertainty estimates. Although Bayesian neural networks (BNNs) attempt to address these shortcomings by treating model weights as random variables, they bring their own challenges. Inference in BNNs is often difficult because the prior choices are not always intuitive, the posterior distributions are complex, and the models have many parameters. In addition, BNNs frequently require repeated forward passes to estimate predictive distributions, resulting in significant computational overhead.

To tackle these issues, recent research has focused on blending the strengths of GPs and neural networks. Deep kernel learning (DKL)~\citep{wilson2016deep} is one such approach, where a neural network transforms the input data into a new feature space, and a GP then operates on these features for prediction. This framework allows the model to optimize both the kernel and the neural network parameters together, using techniques like variational inference or marginal likelihood maximization. Experiments reported in~\cite{wilson2016deep} demonstrate that DKL surpasses standard kernel approaches, such as those using the radial basis function (RBF) kernel, and also performs better than conventional neural networks on various benchmark datasets.

Graph neural networks (GNNs) have become an effective tool for learning from graph structured data (i.e., a structured network of interconnected nodes linked by edges, representing relationships between the nodes), evolving through several stages to address increasingly complex data representation and learning tasks.
The inception of GNNs~\citep{scarselli2008graph} marked a significant shift in how data structured in graphs could be processed by learning algorithms. Graph convolutional networks (GCNs)~\citep{kipf2016semi} introduced the concept of applying convolutional operations directly on graphs. These networks extend the convolutional paradigm to graph data by considering the graph's structure in the convolution operation, allowing for the aggregation of neighbor features through a form of weighted average. This process allows the model to capture the local connectivity and feature patterns in the graph, which has proven effective for node and graph classification tasks. Following GCNs, the development of graph attention networks (GATs)~\citep{velivckovic2017graph} introduced an attention based approach. In GATs, nodes learn to assign different weights to the features of their neighbors, allowing the network to focus on the most relevant information in the local graph structure. GATs have shown significant improvements in various tasks by allowing more nuanced feature aggregation from neighbors, compared to the more uniform aggregation in GCNs. Gated graph sequence networks (GGSNs)~\citep{li2015gated} represent a further evolution in the processing of graph structured data, incorporating elements of sequence modeling into graph networks. By using gated recurrent units~\citep{cho2014learning} or similar mechanisms, GGSNs can model graph dynamics and temporal changes, making them particularly suited for tasks involving sequences of graphs or graphs with evolving structures. This approach has expanded the applicability of GNNs to a broader range of tasks, including those that involve time series data on graphs. 

Despite advances in modeling techniques, many current methods still largely neglect the use of causal knowledge in predictive models. Causal knowledge~\citep{pearl2009causality}, involves understanding and representing cause and effect relationships rather than merely detecting correlations or patterns in data. Most existing data driven and hybrid models primarily identify associations between input features and predicted outcomes, but they often fail to express the underlying causal mechanisms clearly. This lack of causal understanding can significantly limit a model's interpretability, robustness, and ability to generalize, particularly when faced with conditions that differ substantially from those encountered during training. Explicitly integrating causal knowledge into modeling frameworks is, therefore, crucial. It allows models to move beyond simple predictions and actively reason about potential interventions and hypothetical (counterfactual) scenarios. Such an approach provides deeper insights into the underlying physical processes, improving the reliability of predictions. This is especially valuable in complex, dynamic systems such as diesel engines, where accurately distinguishing between mere correlation and genuine causation is essential for accurate diagnostics and effective emission control strategies.

The objective of this paper is to develop and validate a probabilistic model to predict engine-out NOx emissions using Gaussian process regression. Our approach is as follows. We employ three variants of GP models: the first with a standard RBF kernel with input window, the second incorporating a deep kernel using convolutional neural networks (CNNs) to capture temporal dependencies, and the third enriching the deep kernel with a causal graph derived via GCNs.
The causal information derived from the causal graph serves to embed physics informed knowledge into the learning process. 

We have organized our paper as follows. 
We begin by delving into the physical principles underlying the formation of NOx. We then define the problem statement before presenting the methodology. Following this, we verify our approach on an illustrative example. This is followed by the experimental setup, after which we present and discuss our main findings. Finally, we conclude the paper with a summary of the key results and their implications.

This research has been conducted in collaboration with Cummins Inc with the data from a Cummins medium duty diesel engine. In compliance with Cummins policies, all plots in this study have been normalized.

\section{Engine-out $\mathrm{NO}_x$ formation}
\label{sec:engine_out_nox}

The formation of $\mathrm{NO}_x$, consisting of nitric oxide ($\mathrm{NO}$) and nitrogen dioxide ($\mathrm{NO}_2$), in diesel engines involves several mechanisms.  Notably, thermal $\mathrm{NO}$, fuel $\mathrm{NO}$, and prompt $\mathrm{NO}$ play crucial roles~\citep{heywood2018internal}. Prompt $\mathrm{NO}$ predominantly forms under fuel rich conditions and exhibits minimal temperature dependency. Conversely, fuel $\mathrm{NO}$ relies on nitrogenous compounds in the fuel. The dominant process in diesel engines for $\mathrm{NO}$ generation within the combustion chamber is thermal $\mathrm{NO}$, as described by the extended Zeldovich mechanism~\citep{lavoie1970experimental}. This process results from the oxidation of atmospheric nitrogen. The key reactions for thermal $\mathrm{NO}$ are as follows:
\begin{equation*}
\begin{split}
\mathrm{N}_2 + \mathrm{O} \rightarrow \mathrm{NO} + \mathrm{N}, \\
\mathrm{N} + \mathrm{O}_2 \rightarrow \mathrm{NO} + \mathrm{O}, \\
\mathrm{N} + \mathrm{OH} \rightarrow \mathrm{NO} + \mathrm{H}.
\end{split}
\end{equation*}
These above reactions become significant at high temperatures (above $2000\,\mathrm{K}$), and are influenced by the amount of available oxygen ($\mathrm{O}_2$) and the time the gases spend at peak temperatures, especially in lean mixtures~\citep{bowman1975kinetics}. In comparison, $\mathrm{NO}_2$ is mainly produced outside the cylinder, as a result of the partial oxidation of $\mathrm{NO}$ through the reaction:
\begin{equation*}
\mathrm{NO} + \mathrm{HO}_2 \rightarrow \mathrm{NO}_2 + \mathrm{OH}.
\end{equation*}
The final level of $\mathrm{NO}_x$ leaving the engine depends mostly on the temperature of the burnt gases and the oxygen content in the cylinder. These outcomes are affected by several engine operating conditions, including the flow rate of intake air, the rate of fuel injection, and changes in engine speed and load. Consequently, these engine variables are chosen for modeling engine-out $\mathrm{NO}_x$. A widely used technique for reducing $\mathrm{NO}_x$ emissions in diesel engines is exhaust gas recirculation (EGR). EGR works by redirecting a portion of the exhaust back into the combustion chamber. The recirculated gases consist mostly of nitrogen ($\mathrm{N}_2$), carbon dioxide ($\mathrm{CO}_2$), and water vapor ($\mathrm{H}_2\mathrm{O}$), which take the place of some of the fresh intake air. This process lowers both the oxygen content and the peak combustion temperature because the triatomic molecules in EGR have higher specific heat capacities. As a result, less $\mathrm{NO}_x$ is produced~\citep{krishnan2006prediction}.

\section{Problem Statement}
\label{sec:problem_statement}

Despite substantial progress in modeling engine-out NOx, most existing approaches operate within deterministic frameworks that fall short of capturing the various uncertainties present in sensor measurements, engine dynamics, and operating conditions. This limits their robustness and effectiveness for real time diagnostics. Traditional physics based models are computationally demanding and require detailed engine knowledge, while data driven approaches often provide overconfident predictions without uncertainty quantification. Hybrid models partially address these issues, but typically ignore underlying causal mechanisms, reducing interpretability and generalizability.

To overcome these limitations, it is important to develop a new modeling approach that is both reliable and suitable for real world use. First, the model should be able to estimate how certain or uncertain its predictions are, which will make diagnostics more trustworthy and useful. Second, it needs to understand and learn how engine behaviors change over time, since engine processes are dynamic and affected by many factors. Finally, the model should include knowledge about the real causes behind engine operations, so that its predictions are not only accurate but also easier to explain and understand.

\section{Methodology}
\label{sec:methodology}

\subsection{Gaussian Process Regression}
\label{subsec:gpr}

GP is a flexible, probabilistic model often used for regression problems. Instead of assuming a fixed functional form, GPs define a prior directly over possible functions. This prior is completely described by two functions: the mean function $m(\bm{x})$ and the covariance function $k(\bm{x}, \bm{x}')$. The defining property of a GP is that, for any set of input points, the function values follow a joint Gaussian distribution. In GPR, the core idea is to treat the unknown function as a random draw from a GP. The mean function is given by:
\begin{align*}
m(\bm{x}) &= \mathbb{E}[f(\bm{x})],
\end{align*}
while the covariance function is given by:
\begin{align*}
k(\bm{x}, \bm{x}') &= \mathbb{E}[(f(\bm{x}) - m(\bm{x}))(f(\bm{x}') - m(\bm{x}'))].
\end{align*}
In our paper, we use a zero mean function. 
The kernel function measures how similar two points are. One popular choice is the RBF kernel with automatic relevance determination. Assuming $l_d$ as the characteristic length scale corresponding to each dimension $d$, and $D$ denoting the dimensionality, the RBF kernel is defined as:
\begin{equation*}
k_{\alpha}^{\text{rbf}}(\bm{x}, \bm{x}') = \sigma^2 \exp\left(-\sum_{d=1}^D \frac{(x_d - x'_d)^2}{2l_d^2}\right),
\end{equation*}
where $\sigma^2$ represents the signal variance, and $\alpha = (l_1, l_2, \ldots, l_d, \sigma^2)$. The length scale parameters $l_d$ dictate the sensitivity of the kernel to changes along each dimension. A lower value of $l_d$ increases the influence of variations in the corresponding dimension on the covariance.

Consider a dataset $\mathcal{D}$ consisting of $n$ input observations represented by the matrix $X = (\mathbf{x}_1, \dots, \mathbf{x}_n)$, each with dimensionality $D$, and corresponding targets denoted by the vector $\mathbf{y} = (y_1, \dots, y_n)^\top$. Assuming additive Gaussian noise, the likelihood for an individual observation is expressed as:
\begin{equation*}
y_i \mid f(\mathbf{x}_i) \sim \mathcal{N}(y_i; f(\mathbf{x}_i), \sigma_y^2),
\end{equation*}
where $\sigma_y^2$ denotes the variance associated with observational noise. Observations are considered statistically independent. For $n_*$ new test inputs $X_*$, the GP provides the following predictive distribution:
\begin{equation*}
\scalebox{0.85}{$
\begin{gathered}
\mathbf{f}_* \mid X_*, X, \mathbf{y}, \boldsymbol{\theta}, \sigma_y^2 \sim \mathcal{N}\left(\mathbb{E}[\mathbf{f}_*], \mathbb{C}(\mathbf{f}_*)\right), \\
\mathbb{E}[\mathbf{f}_*] = m(X_*) + k_{\theta}(X_*, X)[k_{\theta}(X, X) + \sigma_y^2 I]^{-1}(\mathbf{y} - m(X)), \\
\mathbb{C}[\mathbf{f}_*, \mathbf{f}_*] = k_{\theta}(X_*, X_*) - k_{\theta}(X_*, X)[k_{\theta}(X, X) + \sigma_y^2 I]^{-1}k_{\theta}(X, X_*),
\end{gathered}
$}
\end{equation*}
where,
$$
\mathbf{f}_* = [f_*(\mathbf{x}_1), \ldots, f_*(\mathbf{x}_{n_*})]^\top.
$$
In these equations, $k_{\theta}(X_*, X)$ is the covariance matrix between test and training points, $k_{\theta}(X, X)$ is the covariance among training points, and $m(X_*)$ is the mean at the test inputs.
The model parameters $\theta$ can be determined by maximizing the marginal log likelihood of the observed targets:
\begin{equation*}
\begin{split}
\log p(\mathbf{y} \mid \theta, X) \propto -\left[\underbrace{\mathbf{y}^{\top}(k_{\theta}(X,X) + \sigma_y^2 I)^{-1}\mathbf{y}}_{\text{model fit}} + \right. \\
\left. \underbrace{\log|k_{\theta}(X,X) + \sigma_y^2 I|}_{\text{complexity penalty}}\right].
\end{split}
\end{equation*}

\subsubsection{Deep Kernel}
\label{deep_kernel}

A deep kernel~\citep{wilson2016deep, zinage2024dkl} combines the strengths of neural networks and classical kernel methods to improve a model's ability to capture complicated relationships in data.  The primary function of the neural network component is to perform dimensionality reduction and feature extraction, thereby generating a compact and informative representation suitable for the GP model. Unlike conventional kernels such as the RBF, which inherently assume smoothness and isotropy~\citep{calandra2016manifold}, deep kernels can model complex, irregular, and anisotropic functional relationships. Moreover, the neural network's ability to project high dimensional input spaces onto lower dimensional latent spaces helps mitigate issues associated with the curse of dimensionality. This reduction in dimensionality not only improves computational tractability, but also improves the predictive capability of the GP by focusing on the most informative features of the input data.

Let $\phi(\bm{x};\bm{w})$ denote a nonlinear transformation executed by a neural network characterized by parameters $\bm{w}$, and $k_{\alpha}^{\text{rbf}}(\bm{x}_i, \bm{x}_j')$ represent the standard RBF kernel. The input transformations are then expressed as:
\begin{equation*}
    k_{\theta}^{\text{deep}}(\bm{x}_i, \bm{x}_j')\rightarrow k_{\alpha}^{\text{rbf}}(\phi(\bm{x}_i; \bm{w}), \phi(\bm{x}_j'; \bm{w})),
\end{equation*}
where $\theta = (\alpha, \bm{w})$ collects all the parameters of both the kernel and the neural network. These parameters are learned together by maximizing the log marginal likelihood of the GP model.

\begin{figure}[htbp]
    \centering
    \scalebox{0.7}{
    \includegraphics[width=8.4cm]{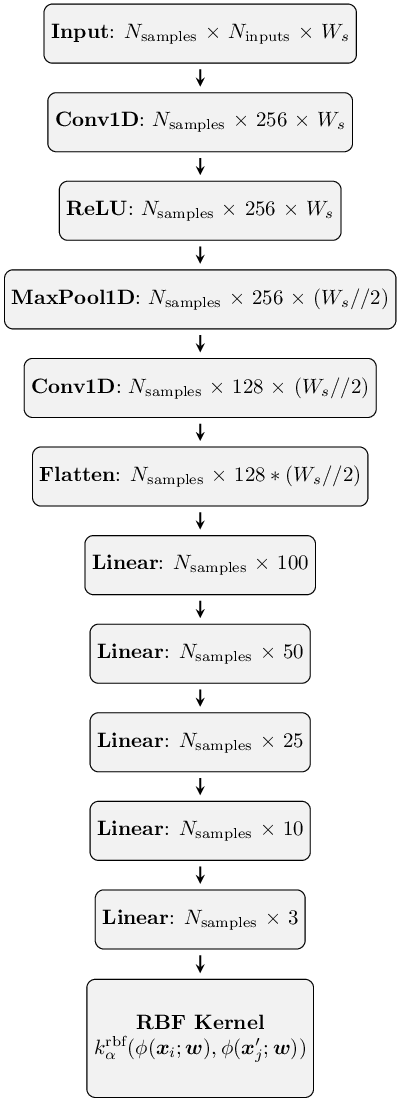}}
    \caption{Deep kernel using CNN.}
    \label{fig:deep_kernel}
\end{figure}

To capture temporal or sequential dependencies effectively, especially when dealing with time series data, a sliding window approach is often used. The sliding window technique involves using a fixed window to extract overlapping segments from the input data, allowing the model to capture both recent and historical patterns in the sequence.
If we consider a sliding input window, we would have the input space dimension to be mapped from a number of input features ($N_\text{inputs}$) $\times$ window size ($W_s$) to a user defined input space before feeding into the RBF kernel. 

In this study, we used CNNs to capture these temporal dependencies as depicted in Fig.\ref{fig:deep_kernel}. CNNs are particularly well suited to capture temporal dependencies due to their translation invariance, local receptive fields, and efficient parameter sharing. Translation invariance allows the model to detect patterns regardless of their position in the sequence, while local receptive fields enable learning of localized temporal dependencies. Shared weights reduce the number of parameters, improving computational efficiency. 

\subsubsection{Deep kernel while incorporating causal information}
\label{subsubsec:deep_kernel_causal}

\begin{figure*}[htbp]
    \centering
    \includegraphics[width = 16.8cm]{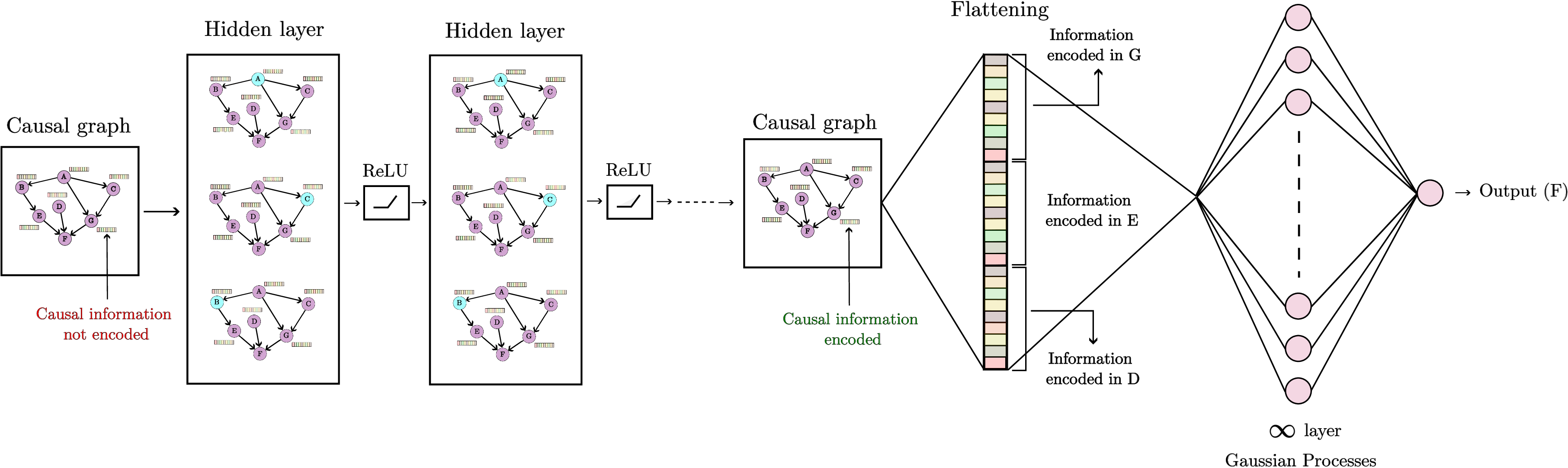}
    \caption{Deep kernel learning while incorporating causal information.}
    \label{fig:dkl_causal}
\end{figure*}

\begin{figure*}[htbp]
    \centering
    \includegraphics[width = 16.8cm]{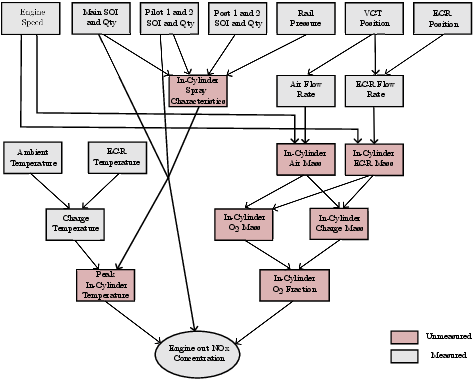}
    \caption{Causal graph for modeling engine-out NOx.}
	\label{fig:causal_graph}
\end{figure*}

GCNs~\citep{kipf2016semi} extend the principles of CNNs to work with graph data. In our approach, we use GCNs to incorporate causal information from the causal graph (Fig. \ref{fig:causal_graph}) into the DKL framework. A graph is denoted as $G = (V, E)$, where $V$ is the set of nodes and $E$ is the set of edges connecting pairs of nodes, with $E \subseteq V \times V$. Each node $v \in V$ is associated with a feature vector $x_v$, which, in our case, contains the time series data for each variable. The collection of all node features forms a matrix $M \in \mathbb{R}^{N \times F}$, where $N$ is the number of nodes and $F$ is the number of features per node. The relationships between nodes are described by the adjacency matrix $A \in \{0, 1\}^{N \times N}$, which encodes the causal graph structure: $A_{ij} = 1$ indicates a direct causal link from node $i$ to node $j$, and $A_{ij} = 0$ otherwise. Let $H^{(l)}$ be the matrix of node features at the $l$-th GCN layer, starting with $H^{(0)} = M$. The trainable weight matrix for layer $l$ is $W^{(l)}$. To incorporate each node’s own features, we add self loops to the adjacency matrix, yielding $\tilde{A} = A + I$, where $I$ is the identity matrix. Assuming $\tilde{D}$ represents the diagonal degree matrix, the standard graph convolution operation in its simplest form can be expressed as:
\begin{equation*}
H^{(l+1)} = \sigma\left(\tilde{D}^{-\frac{1}{2}} \tilde{A}\, \tilde{D}^{-\frac{1}{2}} H^{(l)} W^{(l)}\right),
\end{equation*}
where $\sigma(\cdot)$ is a nonlinear activation function. This operation effectively updates each node's representation by considering its own features and those of its predecessors, as defined by the graph structure.
The normalization with $\tilde{D}^{-\frac{1}{2}} \tilde{A} \tilde{D}^{-\frac{1}{2}}$ ensures that the scale of the feature representations is maintained, preventing vanishing or exploding gradients during training. However, this symmetric normalization is designed for undirected graphs. In our causal context, which is a directed acyclic graph, this formulation would incorrectly allow information to flow from children back to their parents, violating the principles of causality.

So we use a directional GCN variant that strictly respects the causal flow. The message passing mechanism ensures that each node's representation is updated by aggregating information only from its causal parents and itself. Assuming $\hat{A}_{\text{causal}}$ is the normalized propagation matrix that enforces the directed message passing, the propagation rule for the $(l+1)$-th layer is given by:
\begin{equation*}
    H^{(l+1)} = \sigma\left( \hat{A}_{\text{causal}} H^{(l)} W^{(l)} \right).
\end{equation*}
The construction of $\hat{A}_{\text{causal}}$ begins with the transpose of the adjacency matrix $A^T$. The transpose is necessary because in the matrix product $\hat{A}_{\text{causal}} H^{(l)}$, the new features for a node $i$ are a sum over features from nodes $j$. Using $A^T$ ensures that the entry $A^T_{ij}$ is 1 only if $A_{ji}=1$ (i.e., node $j$ is a parent of node $i$), thus guaranteeing correct aggregation from parents to children. We then add self loops to ensure each node retains its own information, yielding the unnormalized propagation matrix $\tilde{A}_{\text{causal}} = A^T + I$. We then apply row normalization to average the incoming messages. Assuming $\tilde{D}_{\text{causal}}$ is the diagonal degree matrix for $\tilde{A}_{\text{causal}}$, the normalized propagation matrix is $\hat{A}_{\text{causal}} = (\tilde{D}_{\text{causal}})^{-1} \tilde{A}_{\text{causal}}$. The use of multiple stacked directed GCN layers allows the model to effectively represent higher order dependencies, thus aiding the learning of rich feature representations that incorporate causal information.

\begin{figure}[htbp]
    \centering
    \scalebox{0.6}{
    \includegraphics[width=8.4cm]{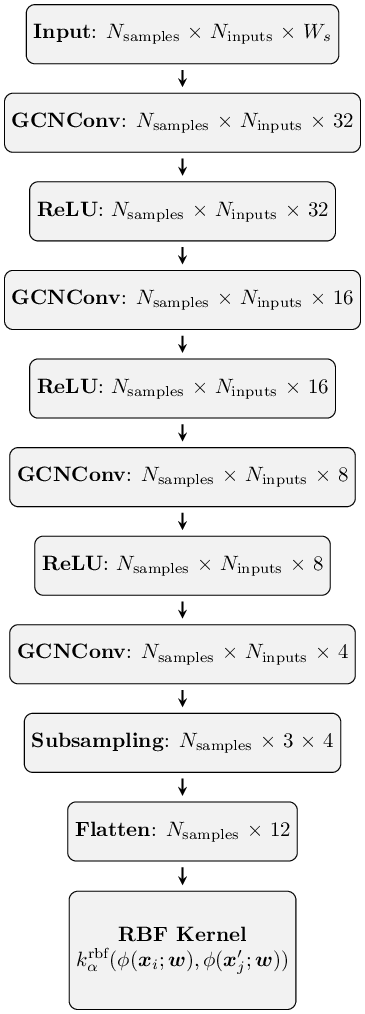}}
    \caption{Deep kernel using directed GCN.}
    \label{fig:deep_kernel_causal}
\end{figure}

The output of the GCN serves as the input to the GP's kernel function. By transforming the input features using the GCN, we provide the GP with rich representations that incorporate both the features and the causal information between the input variables. The RBF kernel is then applied to these transformed features.
Fig. \ref{fig:dkl_causal} shows the schematic of a deep kernel learning while encoding causal information. 
This means that the GCN learns not just the static structure of the graph but also how changes in one part of the graph (e.g., one node or a set of nodes) might influence other parts. 
Once it's trained, even though the output graph may structurally resemble the input, it contains a deeper understanding of the causal relationship within the graph.
Please refer to~\cite{zevcevic2021relating} for further insights into connections between GNNs and structural causal models (SCMs). 
SCMs are frameworks that combine causal graphs with structural equations to model the relationships between variables. Incorporating these structural equations into predictive models via SCMs offers several advantages over traditional statistical or purely correlational models. Traditional models often rely on correlation between variables, which may not reflect true causal relationships. SCMs on the other hand explicitly model causation, ensuring that the influence of each variable on others is accurately represented. Due to this, models based on causation are more likely to generalize well to new unseen data, especially under interventions or distribution shifts. The predictions remain reliable even when the underlying data distribution changes, as causal relationships are invariant to such changes. 

The causal graph for engine-out NOx was constructed based on expert guidance from Cummins, reflecting a physical understanding of how engine-out NOx is caused due to other variables in a diesel compression ignition (CI) engine. Fig. \ref{fig:causal_graph} shows the causal graph for engine-out NOx with the arrows indicating the hypothesized causal link. In Fig. \ref{fig:causal_graph}, certain variables are unmeasured. Consequently, in this study, it is postulated that all causal influences directed towards these unmeasured variables are transmitted without integrating the information of the corresponding variable. One notable exception is the peak in-cylinder temperature, a variable that is critical in influencing the changes in engine-out NOx emissions. However, due to the unavailability of data for peak in-cylinder temperature, it was substituted with turbine inlet temperature, the nearest available analogous variable. Given that the ambient temperature remained constant throughout the data collection phase, all causal relationships involving ambient temperature have been disregarded in this analysis. Also, due to the absence of data for charge temperature, it was substituted with the most closely related measured variable available, namely the intercooler outlet temperature.

The input variables for modeling engine-out NOx were chosen primarily on the basis of the fundamental chemical and thermodynamic principles governing NOx formation.

The inputs we have considered are as follows:
\begin{enumerate}
    \item Turbine inlet temperature
    \item Engine speed
    \item EGR valve actuation
    \item VGT valve actuation
    \item Mass flow rate of EGR
    \item Mass flow rate of air
    \item Fuel rail pressure
    \item Engine brake torque
    \item Main injection timing
    \item Main injection quantity
    \item Pilot 2 injection timing
    \item Pilot 2 injection quantity
    \item Post 1 injection timing
    \item Post 1 injection quantity
    \item EGR system outlet temperature
    \item Intercooler outlet temperature
\end{enumerate}

It is important to note that while directional GCNs incorporate causal graph information into feature representations, they do not perform explicit causal inference or enforce causal
constraints in predictions. Instead, it improves the model's ability to learn complex dependencies by embedding structural information from the causal graph~\citep{thost2021directed}.

\section{Verifying the approach on an illustrative example}
\label{sec:ver_ill_example}

To validate our proposed method, we apply it to a synthetic illustrative example based on a predefined SCM. Consider a system with $k$ input variables and one output variable. We denote the input variables as $x_1, x_2, \dots, x_k$, and the output variable as $y$. Specifically, let $y_t$ represent the target variable at time step $t$.
The SCM for this example defines the following causal relationships:
\begin{equation*}
\begin{aligned}
x_5 &= x_1^2, \\
x_6 &= x_2 \log(1 + |x_4|) + x_3, \\
x_7 &= \sin(x_5) \cos(x_5), \\
x_8 &= x_6^2 + \sqrt{|x_5|}, \\
y   &= \tanh(x_7) + \cos(x_8) + \sin(x_7 - x_8) + x_7 + x_8.
\end{aligned}
\end{equation*}

\begin{figure}[htbp]
\centering
\scalebox{0.7}{
\begin{tikzpicture}

\node[draw, circle, fill=purple!20, minimum size=1.5cm] (c1) at (0, 0) {\Large $x_1$};
\node[draw, circle, fill=purple!20, minimum size=1.5cm] (c2) at (3, 0) {\Large $x_2$};
\node[draw, circle, fill=purple!20, minimum size=1.5cm] (c3) at (6, 0) {\Large $x_3$};
\node[draw, circle, fill=purple!20, minimum size=1.5cm] (c4) at (9, 0) {\Large $x_4$};
\node[draw, circle, fill=green!20, minimum size=1.5cm] (c5) at (6, -3) {\Large $x_6$};
\node[draw, circle, fill=yellow!20, minimum size=1.5cm] (c6) at (6, -6) {\Large $x_8$};
\node[draw, circle, fill=green!20, minimum size=1.5cm] (c7) at (0, -3) {\Large $x_5$};
\node[draw, circle, fill=yellow!20, minimum size=1.5cm] (c8) at (0, -6) {\Large $x_7$};
\node[draw, circle, fill=blue!20, minimum size=1.5cm] (c9) at (3, -9) {\Large $y$};
\draw[->, line width=0.8mm] (c2) -- (c5);
\draw[->, line width=0.8mm] (c3) -- (c5);
\draw[->, line width=0.8mm] (c4) -- (c5);
\draw[->, line width=0.8mm] (c5) -- (c6);
\draw[->, line width=0.8mm] (c1) -- (c7);
\draw[->, line width=0.8mm] (c7) -- (c8);
\draw[->, line width=0.8mm] (c8) -- (c9);
\draw[->, line width=0.8mm] (c6) -- (c9);
\draw[->, line width=0.8mm] (c7) -- (c6);

\end{tikzpicture}}
\caption{Structural causal model (illustrative example).}
\label{fig:SCM_example}
\end{figure}
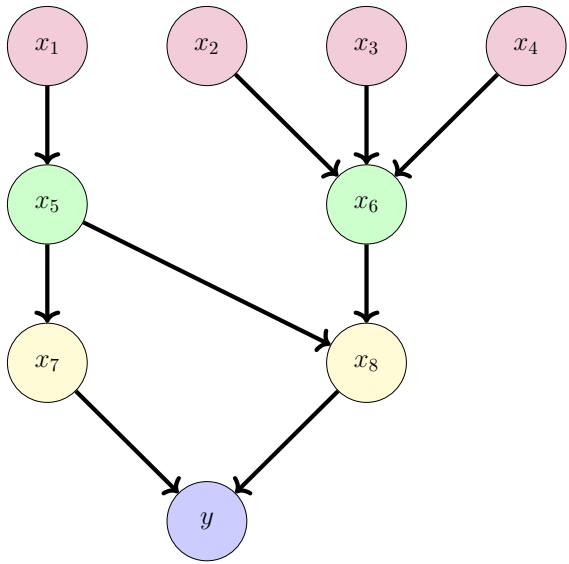

In this model, each equation represents how a child variable is causally influenced by its parent variables. For example, $x_5$ is directly influenced by $x_1$, while $y$ is directly influenced by $x_7$ and $x_8$. This hierarchical and nonlinear structure allows us to assess how well our causal graph-enhanced GP captures the underlying causal dependencies and nonlinear relationships within the data, as depicted in Fig~\ref{fig:SCM_example}. 
To evaluate our approach, we generate synthetic data that adheres to the defined SCM, ensuring the preservation of causal relationships. We begin by sampling the input variables $x_{1,t}, x_{2,t}, x_{3,t}, x_{4,t}$ for each time step $t$ from predefined distributions. Using these sampled inputs, we compute the intermediate variables $x_{5,t}, x_{6,t}, x_{7,t}, x_{8,t}$ according to the SCM equations. The output variable $y_t$ is then defined as follows, with the additive Gaussian noise $\epsilon_t \sim \mathcal{N}(0, \zeta^2)$ to account for measurement uncertainty:
\begin{align*}
y_t &= \tanh(x_{7,t}) + \cos(x_{8,t}) + \sin(x_{7,t} - x_{8,t}) \\
    &\quad + x_{7,t} + x_{8,t} + \epsilon_t.
\end{align*}
Furthermore, we model the measured input variables $x_{i,t,\text{meas}}$ with uncertainty by introducing measurement noise:
\[
x_{i,t,\text{meas}} \mid x_{i,t} \sim \mathcal{N}\left(x_{i,t}, \tau^2\right), \quad i = 1, 2, \dots, k,
\]
where $\tau$ denotes the standard deviation. The GP model is then trained to predict $y_t$ using the inputs while incorporating the information of the causal graph through the deep kernel. We evaluate the performance of our model by comparing it with a standard GP and a GP equipped with a conventional deep kernel using multilayer perceptron (MLP). Additionally, we assess the predictive accuracy of our model against scenarios where the causal information is either partially correct or entirely incorrect, in order to validate the effectiveness of our approach.

To generate synthetic data, we set $\zeta = 0.01$ and $\tau = 0.01$. A total of 10,000 samples are generated, with the first 9000 samples used for training and the last 1000 samples reserved for testing.

We train three GP models: the first with a standard RBF kernel (GP (RBF)), the second with a conventional deep kernel using MLP (GP (DRBF-MLP)), and the third with a deep kernel based on GCN (GP (DRBF-GCN)). To compare the efficacy of our proposed model, we use the root mean squared error (RMSE) as the primary performance metric. Additionally, we assess the model’s performance in scenarios with partially correct or entirely incorrect causal information by using RMSE, mean absolute error (MAE), and the coefficient of determination ($R^2$) as evaluation metrics.

Table~\ref{tab:example_table_1} presents the accuracy of the median predictions of these models, evaluated using RMSE as the number of training samples increases. Notably, the GP (DRBF-GCN) outperforms the other two models, especially in low-data regimes (i.e., when $N=1000$) due to the inductive bias introduced by the causal information encoded in the deep kernel. As the number of training samples increases, the performance of all three models improves, and they converge to similar levels of accuracy.

\begin{table}[htbp]
    \centering
    \caption{Accuracy of the GP models (illustrative example)}
    \label{tab:example_table_1}
    \resizebox{8.5 cm}{!}{
        \begin{tabular}{|c|c|c|c|}
        \hline
            \multirow{2}{*}{\textbf{Models}} & \multicolumn{3}{c|}{\textbf{RMSE (using N training samples)}} \\
            \cline{2-4}
            & \textbf{N = 1000} & \textbf{N = 5000} & \textbf{N = 9000} \\
            \hline
            GP (RBF) & 0.45 & 0.20 & 0.16 \\
            \hline
            GP (DRBF-MLP) & 0.54 & 0.21 & \textbf{0.15} \\
            \hline
            GP (DRBF-GCN) & \textbf{0.18} & \textbf{0.16} & \textbf{0.15} \\
            \hline
        \end{tabular}
    }
\end{table}

Table~\ref{tab:computational_time_table} summarizes the computational training and test times for each of these models when trained and tested on a desktop equipped with an Intel(R) Core(TM) i9-10900K CPU @3.70GHz and an NVIDIA RTX A4000 GPU with 16 GB of GDDR6 memory. Comparable computational efficiency can be observed across methods. Although the GP (DRBF-GCN) model requires slightly more time due to additional graph computations, this modest increase in computational cost is justified by the substantial improvement in accuracy achieved under low data regimes.

\begin{table}[htbp]
    \centering
    \caption{Computational training and test time of GP models (illustrative example)}
    \label{tab:computational_time_table}
    \resizebox{8.5 cm}{!}{
        \begin{tabular}{|c|c|c|}
        \hline
            \textbf{Models} & \textbf{Training time (s) (per epoch)} & \textbf{Test time (s)} \\
            \hline
            GP (RBF) & 0.53 & \textbf{2.1} \\
            \hline
            GP (DRBF-MLP) & \textbf{0.52} & 2.4 \\
            \hline
            GP (DRBF-GCN) & 0.68 & 2.7 \\
            \hline
        \end{tabular}
    }
\end{table}

To further demonstrate that the model effectively encodes causal information, we train three additional GP models: one that encodes the correct causal relationships, one that encodes partially correct causal relationships, and one that encodes incorrect causal relationships. Table~\ref{tab:example_table_2} shows the accuracy of these model's median predictions on the test dataset, evaluated using RMSE, MAE, and $R^2$. We can observe that the highest performance is achieved by the model which encoded the correct causal information with the performance degrading progressively as the encoding of causal information becomes partially correct to completely incorrect. This verifies our proposed approach.

\begin{table}[htbp]
    \centering
    \caption{Accuracy of the GP models with different causal information (illustrative example)}
    \label{tab:example_table_2}
    \resizebox{8.5 cm}{!}{
        \begin{tabular}{|c|c|c|c|}
        \hline
            \textbf{Models} & \textbf{RMSE} & \textbf{MAE} & \textbf{$R^2$} \\
            \hline
            Correct causal information & \textbf{0.15} & \textbf{0.11} & \textbf{0.95} \\
            \hline
            Partially correct causal information & 0.52 & 0.31 & 0.54 \\
            \hline
            Incorrect causal information & 2.31 & 1.98 & -9.68 \\
            \hline
        \end{tabular}
    }
\end{table}

\section{Experimental Setup}
\label{sec:exp_setup}

\subsection{Data Generation}
\label{data_generation}

The experimental data used in this research was obtained from a Cummins mid range multi pulse B6.7L inline 6-cylinder diesel CI engine equipped with high pressure common rail fuel injection system, a high pressure EGR system, and a VGT. This engine was tested at Cummins Technical Center in Columbus, IN. The data is a mixture of lab grade instrumentation and on-engine production sensors. Key specifications of the engine from which the experimental data were derived are presented in Table \ref{tab:engine-specs}.

\begin{table}[htbp]
\centering
\caption{Specifications of the engine}
\label{tab:engine-specs}
\begin{tabular}{|c|c|} 
\hline
\textbf{Specification}            & \textbf{Details}                         \\ \hline
Engine type              & Cummins B6.7L CI engine                       \\
\hline
Horsepower                  & 200-325 hp (149--242 kW)                \\
\hline
Peak torque                & 520-750 lb-ft (705-1017 Nm)           \\
\hline
Governed speed                    & 2600 rpm                                 \\
\hline
Clutch engagement torque          & 400 lb-ft (542 Nm)                      \\
\hline
Number of cylinders               & 6                                        \\
\hline
Engine weight (dry)               & 1150 lb (522 kg)                        \\
\hline
Fuel system                       & High pressure common rail         \\
\hline
Turbocharger                      & VGT     \\
\hline
Emissions control                 & High pressure EGR \\
\hline
Certification                     & EPA 2021                                 \\ \hline
\end{tabular}
\end{table}

The performance of the trained models is then tested on six different validation datasets that were collected on various duty cycles by intentionally running at varying engine-out NOx levels. In our discourse, we consider the experimental data to be the ground truth and the modeling results are compared with it.

\subsection{Data Normalization}
\label{data_normalization}

Prior to training the GP model, we normalized all datasets using the empirical cumulative distribution function (ECDF) method (also known as the quantile transform method) \citep{peterson2020ordered}. This normalization converts the data into a uniform/Gaussian distribution. In our study, we specifically transformed the data into a uniform distribution. This approach is especially beneficial when the original distribution is unknown or does not conform to the Gaussian distribution often assumed by many machine learning algorithms. Additionally, the ECDF approach improves the robustness to outliers by ranking data points instead of directly scaling their values.

\subsection{Metrics}
\label{metrics}

In order to assess the ability of the GP models to accurately model NOx, we employed several quantitative metrics. These include the RMSE, the 90th, 95th, and 98th percentiles of the absolute errors in NOx emissions, which provide a comprehensive view of the error distribution. RMSE was selected as the primary statistic due to its sensitivity to large deviations, which aligns with the critical importance of accurately modeling peak emission events in combustion systems. Unlike MAE, which treats all errors uniformly, RMSE disproportionately penalizes larger errors, a particularly relevant factor in emissions modeling, where infrequent but extreme deviations can have regulatory implications.
Similarly, $R^2$ while informative, is less interpretable in scenarios involving nonlinear, heteroskedastic data distributions and may offer limited insight into absolute prediction quality.
We also included the 90th, 95th, and 98th percentiles of absolute errors to better understand how the model performs under the most challenging conditions. These values tell us the error below which 90$\%$, 95$\%$, and 98$\%$ of the predictions fall. In other words, they show how large the biggest errors are for a small portion of the predictions. This is especially important in emissions modeling, where even a few large errors can lead to violations of strict environmental regulations. By looking at these high percentiles, we can better judge whether the model is reliable in situations where accuracy matters the most.
Additionally, to examine the consistency of NOx emissions under identical input conditions, we used the Quantile-Quantile (Q-Q) plot, which provides a graphical representation of the empirical distribution of the model's output compared to a theoretical distribution, thereby allowing a thorough assessment of the model's ability to replicate the observed NOx distribution under repeated experiments.

\section{Results}
\label{sec:results}

The GP models and deep kernel using GCN are implemented using the GPyTorch~\citep{gardner2018gpytorch} and PyTorch Geometric~\citep{fey2019fast} libraries, respectively, each using a PyTorch backend. The negative exact marginal log likelihood serves as the loss function, while optimization is performed using the Adam optimizer~\citep{kingma2014adam}, with tuned hyperparameters.
To prevent the models from overfitting, we used early stopping with a patience of 50 epochs.

\begin{figure*}[htbp]
    \centering
    \includegraphics[width = 16.8cm]{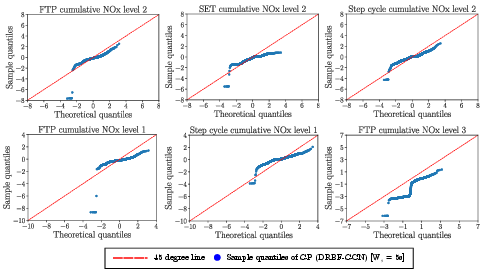}
    \caption{Quantile-Quantile plots (GP(DRBF-GCN)[$W_s = 5s$]).}
    \label{fig:qq_plots}
\end{figure*}

In our comprehensive analysis of various GP models built for predicting engine-out NOx, we were provided with the readings from an ECM virtual sensor provided by Cummins for the purpose of comparison. 
This comparative study revealed distinct trends and performance nuances in our models. Firstly, there is a clear indication that increasing the complexity of GP models improves their predictive performance. This is particularly noticeable when comparing the performance of standard GP (RBF) models with different input window sizes against relatively complex models such as GP (Deep RBF) with CNN or GCN. These complex models often outperform both the standard GP models and the ECM virtual sensor provided by Cummins, especially in scenarios with cumulative NOx level 2 (as seen in Tables \ref{tab:val_1_table} and \ref{tab:val_2_table}). However, this trend is not universal across all test cases, suggesting that while complexity contributes to performance, it is not the sole determinant. We can see that GP (DRBF-GCN) [$W_s = 5s$] performs relatively better than all other models for validation 1, 2, and 6 (see Tables \ref{tab:val_1_table}, \ref{tab:val_2_table} and \ref{tab:val_6_table}). However, we observe a relatively better performance of more simpler models for cumulative NOx level 1 scenarios (as seen in Tables \ref{tab:val_4_table} and \ref{tab:val_5_table}).

\begin{table}[htbp]
    \centering
    \caption{Validation 1 (FTP cumulative NOx level 2) (in ppm)}
    \label{tab:val_1_table}
    \resizebox{8.5 cm}{!}{\begin{tabular}{|c|c|c|c|c|}
    \hline
         \multirow{2}{*}{\textbf{Models}} & \multirow{2}{*}{\textbf{RMSE}} & \multicolumn{3}{c|}{\textbf{NOx Error Percentiles}} \\
         \cline{3-5}
         & & \textbf{90th} & \textbf{95th} & \textbf{98th} \\
         \hline
         GP (RBF) [$W_s = 1s$]&  61.39 & 75.86 &  111.10& 174.29\\
         \hline
         GP (RBF) [$W_s = 2s$]& 63.30 & 68.81 & 118.17 & 193.81\\
         \hline
         GP (RBF) [$W_s = 3s$]&  61.68 & 73.78& 113.68 &187.26\\
         \hline
         GP (RBF) [$W_s = 4s$]&  63.08 & 80.54 & 117.73 & 174.86\\
         \hline
         GP (RBF) [$W_s = 5s$]&  66.58 & 77.56& 120.20 & 203.28\\
         \hline
         GP (DRBF-CNN) [$W_s = 5s$]& 61.46  & 80.18 & 120.83 & 176.45\\
         \hline
         GP (DRBF-GCN) [$W_s = 5s$]& \textbf{60.73}  & 74.58 & 110.33 & 170.36\\
         \hline
         ECM virtual sensor& 102.53  & 142.88 & 189.14 & 361.77\\
         \hline
    \end{tabular}}
\end{table}

\begin{table}[htbp]
    \centering
    \caption{Validation 2 (SET cumulative NOx level 2) (in ppm)}
    \label{tab:val_2_table}
    \resizebox{8.5 cm}{!}{\begin{tabular}{|c|c|c|c|c|}
    \hline
         \multirow{2}{*}{\textbf{Models}} & \multirow{2}{*}{\textbf{RMSE}} & \multicolumn{3}{c|}{\textbf{NOx Error Percentiles}} \\
         \cline{3-5}
         & & \textbf{90th} & \textbf{95th} & \textbf{98th} \\
         \hline
         GP (RBF) [$W_s = 1s$]&  46.20 & 81.78 & 97.09 & 105.62\\
         \hline
         GP (RBF) [$W_s = 2s$]&  46.55 & 64.25 & 69.62 & 91.48\\
         \hline
         GP (RBF) [$W_s = 3s$]& 42.87  & 74.09 & 82.32 & 86.66\\
         \hline
         GP (RBF) [$W_s = 4s$]&  44.44 & 74.90 & 83.69 & 86.88\\
         \hline
         GP (RBF) [$W_s = 5s$]&  45.78 & 75.07 & 82.55 & 85.91\\
         \hline
         GP (DRBF-CNN) [$W_s = 5s$]&  50.16 & 83.66 & 92.35 & 97.72\\
         \hline
         GP (DRBF-GCN) [$W_s = 5s$]& \textbf{42.86}  & 74.18 & 82.45 & 86.95\\
         \hline
         ECM virtual sensor&  58.59 & 105.38 & 111.58 & 140.42\\
         \hline
    \end{tabular}}
\end{table}

\begin{table}[htbp]
    \centering
    \caption{Validation 3 (Step cycle cumulative NOx level 2) (in ppm)}
    \label{tab:val_3_table}
    \resizebox{8.5 cm}{!}{\begin{tabular}{|c|c|c|c|c|}
    \hline
         \multirow{2}{*}{\textbf{Models}} & \multirow{2}{*}{\textbf{RMSE}} & \multicolumn{3}{c|}{\textbf{NOx Error Percentiles}} \\
         \cline{3-5}
         & & \textbf{90th} & \textbf{95th} & \textbf{98th} \\
         \hline
         GP (RBF) [$W_s = 1s$]&  117.62 & 126.26 &  185.99& 313.48\\
         \hline
         GP (RBF) [$W_s = 2s$]&  \textbf{73.70} & 98.19 &141.58  & 196.38\\
         \hline
         GP (RBF) [$W_s = 3s$]&  82.45 & 132.19 & 175.77 & 250.93\\
         \hline
         GP (RBF) [$W_s = 4s$]&  90.24 & 132.87 & 183.10 & 268.93\\
         \hline
         GP (RBF) [$W_s = 5s$]&  101.43 & 147.34 & 197.26 &294.08 \\
         \hline
         GP (DRBF-CNN) [$W_s = 5s$]&  81.33 & 101.54 & 153.63 & 254.07\\
         \hline
         GP (DRBF-GCN) [$W_s = 5s$]& 83.54  & 103.82 & 158.26 & 241.45\\
         \hline
         ECM virtual sensor& 216.60  &  349.67&  561.70& 758.07\\
         \hline
    \end{tabular}}
\end{table}

\begin{table}[htbp]
    \centering
    \caption{Validation 4 (FTP cumulative NOx level 1) (in ppm)}
    \label{tab:val_4_table}
    \resizebox{8.5 cm}{!}{\begin{tabular}{|c|c|c|c|c|}
    \hline
         \multirow{2}{*}{\textbf{Models}} & \multirow{2}{*}{\textbf{RMSE}} & \multicolumn{3}{c|}{\textbf{NOx Error Percentiles}} \\
         \cline{3-5}
         & & \textbf{90th} & \textbf{95th} & \textbf{98th} \\
         \hline
         GP (RBF) [$W_s = 1s$]& 89.56  & 100.05 &  132.69& 199.25\\
         \hline
         GP (RBF) [$W_s = 2s$]&  80.38 & 103.20 & 134.09 & 169.11\\
         \hline
         GP (RBF) [$W_s = 3s$]&  77.94 & 94.85 & 124.80 & 171.09\\
         \hline
         GP (RBF) [$W_s = 4s$]&  \textbf{77.37} & 88.78 & 116.60 & 148.06\\
         \hline
         GP (RBF) [$W_s = 5s$]& 83.37  & 98.78 & 125.67 & 152.97\\
         \hline
         GP (DRBF-CNN) [$W_s = 5s$]& 126.98  & 209.13 & 291.76 & 334.32\\
         \hline
         GP (DRBF-GCN) [$W_s = 5s$]& 93.71  & 110.48 & 143.31 & 205.34\\
         \hline
         ECM virtual sensor&  152.79 & 225.43 & 287.62 & 400.98\\
         \hline
    \end{tabular}}
\end{table}

\begin{table}[htbp]
    \centering
    \caption{Validation 5 (Step cycle cumulative NOx level 1) (in ppm)}
    \label{tab:val_5_table}
    \resizebox{8.5 cm}{!}{\begin{tabular}{|c|c|c|c|c|}
    \hline
         \multirow{2}{*}{\textbf{Models}} & \multirow{2}{*}{\textbf{RMSE}} & \multicolumn{3}{c|}{\textbf{NOx Error Percentiles}} \\
         \cline{3-5}
         & & \textbf{90th} & \textbf{95th} & \textbf{98th} \\
         \hline
         GP (RBF) [$W_s = 1s$]& 123.90  & 190.46 & 247.38 & 392.68\\
         \hline
         GP (RBF) [$W_s = 2s$]&  85.35 & 132.38 & 177.28 & 256.68\\
         \hline
         GP (RBF) [$W_s = 3s$]&  \textbf{84.83} & 130.67 & 189.10 & 277.71\\
         \hline
         GP (RBF) [$W_s = 4s$]&  92.76 & 147.26 & 212.70 & 313.76\\
         \hline
         GP (RBF) [$W_s = 5s$]&  96.39 & 151.64 & 229.76 & 313.40\\
         \hline
         GP (DRBF-CNN) [$W_s = 5s$]&  89.66 & 138.24 & 186.94 & 265.52\\
         \hline
         GP (DRBF-GCN) [$W_s = 5s$]&  91.58 & 143.65 & 188.63 & 265.20\\
         \hline
         ECM virtual sensor&  213.85 & 324.20 & 510.12 & 711.17\\
         \hline
    \end{tabular}}
\end{table}

\begin{table}[htbp]
    \centering
    \caption{Validation 6 (FTP cumulative NOx level 3) (in ppm)}
    \label{tab:val_6_table}
    \resizebox{8.5 cm}{!}{\begin{tabular}{|c|c|c|c|c|}
    \hline
         \multirow{2}{*}{\textbf{Models}} & \multirow{2}{*}{\textbf{RMSE}} & \multicolumn{3}{c|}{\textbf{NOx Error Percentiles}} \\
         \cline{3-5}
         & & \textbf{90th} & \textbf{95th} & \textbf{98th} \\
         \hline
         GP (RBF) [$W_s = 1s$]&  438.44 & 691.64 & 703.77 & 731.37\\
         \hline
         GP (RBF) [$W_s = 2s$]&  473.84 & 791.03 & 819.27 & 848.78\\
         \hline
         GP (RBF) [$W_s = 3s$]&  345.26 & 569.79 & 584.68 & 616.90\\
         \hline
         GP (RBF) [$W_s = 4s$]&  289.50 & 468.79 & 481.60 & 515.96\\
         \hline
         GP (RBF) [$W_s = 5s$]& 285.97  & 467.26 & 484.28 & 527.75\\
         \hline
         GP (DRBF-CNN) [$W_s = 5s$]&  257.10 & 399.96 & 427.23 & 455.54\\
         \hline
         GP (DRBF-GCN) [$W_s = 5s$]& 138.18  & 123.43 & 183.87 & 380.44\\
         \hline
         ECM virtual sensor&  \textbf{107.55} & 108.35 & 169.16 & 349.40\\
         \hline
    \end{tabular}}
\end{table}

\begin{figure}[htbp]
    \centering
    \includegraphics[width=8.4cm]{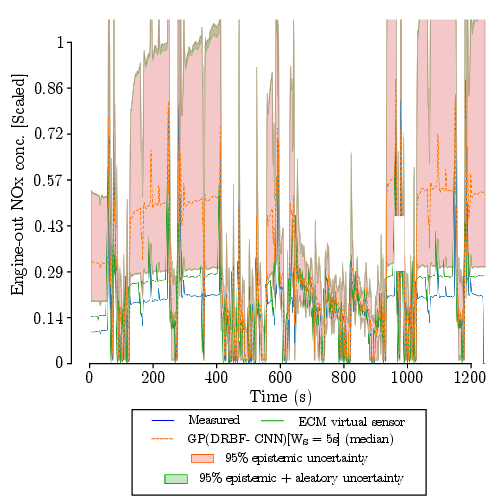}
    \caption{GP(DRBF-CNN)[$W_s = 5s$] predictions on FTP with cumulative NOx level 3.}
    \label{fig:val_6_cnn}
\end{figure}

\begin{figure}[htbp]
    \centering
    \includegraphics[width=8.4cm]{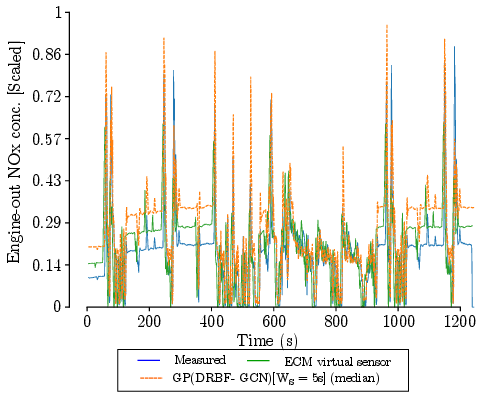}
    \caption{GP(DRBF-GCN)[$W_s = 5s$] predictions on FTP with cumulative NOx level 3.}
    \label{fig:val_6_gnn}
\end{figure}

It is important to differentiate between epistemic and aleatory uncertainties in this context. Epistemic uncertainty originates from insufficient knowledge or data within the model, while aleatory uncertainty is due to the inherent variability in the system under study. In the cumulative NOx level 3 scenario of validation 6, all GP models including the more complex ones, underperform significantly compared to the ECM sensor. The ECM sensor's relatively better performance in this scenario can be linked to its training on a different dataset, likely encompassing a wider representation of NOx conditions with cumulative NOx level 3. This highlights a potential gap in our training dataset for scenarios with cumulative NOx level 3, particularly in terms of epistemic uncertainty. Fig. \ref{fig:val_6_cnn} further supports this, showing considerable epistemic uncertainty as compared to aleatory uncertainty in the GP (DRBF-CNN) [$W_s = 5$s] model predictions for FTP with cumulative NOx level 3, suggesting encounters with scenarios not adequately captured in the training phase.  Due to encoding of causal relationships in the deep kernel for GP (DRBF-GCN), we can observe a relative decrease in the offset between the median prediction and the measured data (Fig. \ref{fig:val_6_gnn}) as compared to GP (DRBF-CNN). Although incorporating physical laws did not help us remove the offset completely, it was at least able to decrease this offset compared to pure regression models. Please note that all the values presented in the tables are expressed in parts per million (ppm).

To analyze the variability of NOx conditioned on the same input, Fig. \ref{fig:qq_plots} shows the Q-Q plots between the sample quantiles of the GP (DRBF-GCN) [$W_s = 5$s] model (in the scaled version) and the theoretical quantiles of standard normal for all validation datasets. In an ideal case these distributions should be the same i.e., all the points must lie on the 45 degree dotted red line. We can see that except for FTP with cumulative NOx level 3, the intermediate NOx values closely follow the standard normal. There is a reasonable difference observed between the sample and theoretical quantiles for extreme low and high NOx values. This is indicative that our model is very sensitive to the changes in input values for predicting extreme low or high NOx values. Due to the gap between validation 6 and the training datasets, we can observe reasonable deviations between these two quantiles for FTP with cumulative NOx level 3.  We could not compare these results with the ECM virtual sensor as it was not a probabilistic model.

\section{Conclusions}
\label{sec:conclusions}

This paper developed and validated a set of probabilistic models for predicting engine-out NOx using GPR. These models were compared against a virtual ECM sensor, providing a robust framework for assessing their predictive performance under different operating conditions. We systematically increased the complexity of the model in three stages: 
\begin{itemize}
    \item First, by incorporating memory through varying input window sizes.
    \item Second, by using CNN within the deep kernel framework to improve the model’s ability to learn complex temporal patterns.
    \item Finally, by using GCN to incorporate causal information, embedding knowledge informed by physics into the learning process.
\end{itemize}

Key findings indicate that increased model complexity improved prediction accuracy in cumulative NOx level 2 scenarios, with the GP (DRBF-GCN) model consistently outperforming simpler models and the ECM sensor. The incorporation of causal information in the GP (DRBF-GCN) model reduced the offset between predictions and actual measurements for the FTP cycle with cumulative NOx level 3, although the performance in extreme NOx cases varied. Simpler models performed better in scenarios with cumulative NOx level 1, and NOx predictions (more specifically, the FTP cycle with cumulative NOx level 3) suffered from high epistemic uncertainty due to insufficient training data. 

Although the GP (DRBF-GCN) model demonstrates superior performance in encoding and leveraging causal information for NOx prediction, several limitations must be considered. 
\begin{itemize}
    \item The efficacy of GP (DRBF-GCN) is heavily based on the accuracy of the encoded causal relationships. As illustrated in Table~\ref{tab:example_table_2}, the performance of the model significantly degrades when causal information is partially correct or incorrect, highlighting its sensitivity to the quality of causal encoding. This dependence requires precise identification and integration of causal factors, which may not always be feasible in complex engine systems. 
    \item While GCNs incorporate causal graph information into feature representation, they impose only a soft constraint. This approach makes GCNs relatively less sensitive to partially correct/incorrect causal information.
    \item Additionally, given a sufficiently large dataset, a standard GP can learn all relevant feature interactions. Under such conditions, the standard GP sometimes may even outperform a GCN, especially if the GCN architecture is excessively rigid or is inadequately aligned with the underlying structure of the data.
\end{itemize}

Future work will address the current limitations of the modeling framework, specifically its inability to transfer effectively from one engine to another due to sensor biases. To improve the generalizability and applicability of these predictive models, we plan to develop a framework capable of inferring/estimating these sensor biases for each engine, allowing corrections to be made dynamically as models transition between engines. By incorporating such bias-correction mechanisms, the need for extensive retraining of the model is significantly reduced, allowing the use of a pre-trained model with minimal adjustments.
This approach will ensure consistent and accurate predictions across different engines, improving both the robustness and efficiency of the modeling process.

\paragraph{\normalfont{\textbf{Authors' contributions}}}
\hfill \\
\textbf{Shrenik Zinage:} Methodology, Software, Validation, Visualization, Writing - original draft. \textbf{Ilias Bilionis:} Funding acquisition, Methodology, Validation, Writing - review and editing. \textbf{Peter Meckl:} Funding acquisition, Writing - review and editing.

\paragraph{\normalfont{\textbf{Acknowledgement}}}
\hfill \\
The authors extend sincere gratitude to Akash Desai from Cummins Inc. for his invaluable feedback and guidance throughout this research. Additionally, heartfelt thanks are due to Dr. Lisa Farrell and Clay Arnett from Cummins Inc., not only for sponsoring this research but also for their crucial technical input and the provision of experimental data essential for conducting the simulations.

\paragraph{\normalfont{\textbf{Declaration of conflicting interests}}}
\hfill \\
The author(s) declared no potential conflicts of interest with respect to the research, authorship, and/or publication of this article.

\paragraph{\normalfont{\textbf{Funding}}}
\hfill \\
This work has been funded by Cummins Inc under grant number 00099056.

\bibliography{refs}

\end{document}